\def\year{2020}\relax
\documentclass[letterpaper]{article} 
\usepackage{aaai20}  
\usepackage{times}  
\usepackage{helvet} 
\usepackage{courier}  
\usepackage[hyphens]{url}  
\usepackage{graphicx} 
\usepackage{graphicx}  
\usepackage{amsmath}
\usepackage{amssymb}
\usepackage{algorithm, algorithmic}
\usepackage{booktabs}
\usepackage{multirow}
\usepackage{makecell,rotating}
\usepackage{subfigure}
\usepackage{url}
\usepackage{breakurl}
\newcommand{\para}[1]{\smallskip\noindent\textbf{#1}}

\newcommand{\wm}{\normalsize{\texttt{WATERMARK} }}
\newcommand{\tabincell}[2]{\begin{tabular}{@{}#1@{}}#2\end{tabular}}

\title{Attacking Optical Character Recognition (OCR) Systems\\ with Adversarial Watermarks}
\author{Lu Chen\textsuperscript{\rm 1} and \Large \textbf{Wei Xu\textsuperscript{\rm 1}}  \\  
\textsuperscript{\rm 1}Institute for Interdisciplinary Information Sciences, Tsinghua University, Beijing, China \\ 
lchen17@mails.tsinghua.edu.cn \\
weixu@tsinghua.edu.cn 
}
 
\begin{document}

\maketitle

\begin{abstract}
Optical character recognition (OCR) is widely applied in real applications serving as a key preprocessing tool. The adoption of deep neural network (DNN) in OCR results in the vulnerability against adversarial examples which are crafted to mislead the output of the threat model. Different from vanilla colorful images, images of printed text have clear backgrounds usually. However, adversarial examples generated by most of the existing adversarial attacks are unnatural and pollute the background severely. To address this issue, we propose a watermark attack method to produce natural distortion that is in the disguise of watermarks and evade human eyes' detection. Experimental results show that watermark attacks can yield a set of natural adversarial examples attached with watermarks and attain similar attack performance to the state-of-the-art methods in different attack scenarios.
\end{abstract}

\section{Introduction}

Optical Character Recognition (OCR) is a widely adopted application for conversing printed or handwritten images to text, which becomes a critical preprocessing component in text analysis pipelines, such as document retrieval and summarization. OCR has been significantly improved in recent years thanks to the wide adoption of the deep neural network (DNN), and thus deployed in many critical applications where OCR's quality is vital. For example, photo-based ID recognition depends on OCR's quality to automatically structure information into databases, and automatic trading sometimes relies on OCR to read certain news articles for determining the sentiment of news. 

Unfortunately, OCR also inherits all counter-intuitive security problems of the DNNs. Especially, the OCR model is also vulnerable to \emph{adversarial examples}, which are crafted by making human-imperceptible perturbations on original images with the intent of misleading the model.  The wide adoption of OCR in real pipelines gives more incentives for adversaries to game the OCR,  such as causing fake ID information, incorrect readings of metrics or instructions, etc. Figure~\ref{fig:license_sample} and \ref{fig:report_sample} in the evaluation section illustrate two real-world examples with attacking the ID number and financial report number.  This paper provides a preliminary study on the possibility of OCR attacks.



Many prior works~\cite{nguyen2015deep,goodfellow2014explaining,papernot2016limitations,szegedy2013intriguing} have shown that changing the prediction of DNNs is practicable by applying carefully-designed perturbations (usually background noise) to original images, in traditional image classification tasks. Recent projects, Adversarial Patch \cite{brown2017adversarial} and LaVAN \cite{karmon2018lavan}, introduced the adversarial patch attack, which puts visible perturbations confined to a small region or location in the image.

However, these methods are not directly applicable to OCR attacks for the following three reasons:

First, the input image to OCR is on a white paper with a spotless background.  Thus any perturbation added by existing attacks will appear so obvious to human readers that it will cause suspicion. 

Second, in complex languages like Chinese, there are many characters (e.g., the dataset we use contains 5,989 unique characters).  If an adversary wants to perform a \emph{targeted} attack, i.e., changing one character to another specific one (target) in a sentence and meanwhile resulting in semantically meaningful recognition results, it requires a large number of perturbations that are too obvious to hide.  

Third, instead of classifying characters individually, the modern OCR model is an end-to-end neural network, inputing a variable-sized image and outputting sequences of labels.  In other words, it works on feeding images line by line. It is usually called the \textit{sequential labeling} task, which is relatively harder to be attacked than the image classification task. It is insufficient to just add perturbations to a single character. Instead, the perturbations are required to span multiple characters.  Also, as the OCR model is end-to-end, the internal feature representations rely on nearby characters (contexts). Thus the perturbations of attacking a single character are designed given its contexts.



In this preliminary study, we propose a new attack method, {\wm} attack, against modern OCR models. Watermarks are images or patterns commonly used in documents as background to identify things, such as marking a document proprietary, secret, urgent, or even simply as decoration.  Similar to watermarks, in Asian countries, documents often contain stamps to verify their authority.  Human eyes are so used to these watermarks and ignore them.  In this paper, we generate natural watermark-style perturbations. That is, we limit all perturbations within a small region of the image, i.e., a watermark.  Given the bound, we minimize the perturbation level. In comparison, classic adversarial examples spread noise all over the image. Our approach is similar to the patch-based attacks \cite{brown2017adversarial,karmon2018lavan}. Different from that patches absolutely cover part of the images, watermarks do not hinder text's readability and thus look more natural. \cite{heng2018harmonic} generated disguising perturbations as shadows, exposure problems or color problems. And \cite{hanwei2019smooth} generated smooth noise by Laplacian smoothing. But none of these solve the clear background challenge for OCR. 

We focus on \emph{white-box, targeted attack} in this paper. That is, we assume adversaries have perfect knowledge of the DNN architecture and parameters (white-box model) and aim to generate specific recognition results (targeted attack).  Given that many real OCR softwares are based on similar open-source OCR DNN models, we believe the white-box model, in addition to being a starting point, also has real-world applicability.  

As a consequence, the {\wm} attack is an adversarial attack on the OCR model. The {\wm} attack attaches natural watermark-style noise, tricks the OCR model into outputting specific recognition results, and preserves the readability of adversarial images at the same time. To some extent, the {\wm} attack solves the clear background problem.

As an evaluation, we performed the {\wm} attack on a state-of-the-art open-source OCR model using DenseNet + CTC neural network architectures for the Chinese language. We used a dataset with 3.64 million images and 5,989 unique characters. With 158 pairs of original-target, we show that the {\wm} attack can generate quite human-eye friendly adversarial samples with a high probability of success. Some of {\wm} adversarial examples even work on Tesseract OCR~\cite{Tesseract} in a black-box manner.

Even more, we applied our model to real-world scenarios. In Figure \ref{fig:report_sample}, we employed the {\wm} method to a page of an annual report of a listed Chinese company and changed the semantics, in the meantime, the image looks natural to human readers.


The contributions of this paper include:
1) We propose the {\wm} adversarial attack to generate natural-to-human watermark-style perturbations, targeting DNN-based OCR systems. We also demonstrated a method to hide perturbations that human eyes are accustomed to, in a watermark-bounded region .  
2) Using difficult OCR cases (Chinese), we demonstrated the success rate of {\wm} attacks comparing to existing ones.  


\section{Background and Related Work}

\subsection{Optical Character Recognition (OCR)}

Generally, the OCR pipeline, as shown in Figure \ref{fig:flowchart}, begins with line segmentation, which includes page layout analysis for locating the position of each line, de-skewing the image, and segmenting the input image into line images. After preprocessing line images, such as rescaling and normalizing, such images are fed into the recognition model, which outputs recognition results. 

There are two types of OCR models.
1) Character-based models are the classic way~\cite{smith2007overview}. Such a recognition model segments the image into per-character sub-images and classifies each sub-image into the most likely recognition result. Obviously, its performance heavily relies on the character segmentation. 
2) End-to-end models are a segmentation-free approach. It recognizes entire sequences of characters in a variable-sized image. \cite{bengio1995lerec,espana2011improving} adopted sequential models. \cite{breuel2013high,wang2012end} utilize DNNs as the feature extractor for the end-to-end sequential recognition. Sequential DNN models \cite{graves2006connectionist} introduced a segmentation-free approach, connectionist temporal classification(CTC), which allows variable-sized input images and output results.

In end-to-end models, \emph{sequence labeling} is a task that assigns a sequence of discrete labels to variable-sized sequential input data. In our case, the input is a variable-size image $\boldsymbol{x}$ and the output is a sequence of characters $\boldsymbol{t} = [t_1, t_2, ..., t_N]$, $t_i \in \mathcal{T}$ from predefined character set $\mathcal{T}$. 

\para{Connectionist Temporal Classification (CTC). }  CTC is an alignment-free method for training DNNs on the sequential labeling task, which provides a kind of loss enabling us to recognize sequences without explicit segmentation while training DNNs. Therefore, many state-of-the-art OCR models use CTC as the model's loss function. Given the input image $\boldsymbol{x}$, let $f(\boldsymbol{x}) = \boldsymbol{y} = [y_1, y_2, ..., y_M]$ be the sequence of model $f$'s outputs, where $M \geq N$ and $y_i \in [0, 1]^{|\mathcal{T}|}$ is the probability distribution over the character set $\mathcal{T}$ in observing label $i$.

CTC requires calculating the likelihood $p(\boldsymbol{t}|\boldsymbol{x})$, which is barely directly measured from the model's probability distribution $f(\boldsymbol{x})$ and the target sequence $\boldsymbol{t}$. To settle this, CTC uses a valid alignment $\boldsymbol{a}=[a_1, a_2, ..., a_M]$ of $\boldsymbol{t}$, $a_i \in \mathcal{T} \cup \{blank\}$, where the target sequence $\boldsymbol{t}$ can be obtained by removing all blanks and sequential duplicate characters (e.g. both [a, --, a, b, --] and [--, a, a, --, --, a, b, b] are valid alignments of [a, a, b]). The likelihood  $p(\boldsymbol{t}|\boldsymbol{x})$ is to sum up the probability of all possible valid alignments denoted as $A$.
\begin{align}
    p(\boldsymbol{t}|\boldsymbol{x}) = \sum_{\boldsymbol{a}\in A} \prod_{i=1}^M p(a_i | \boldsymbol{x}) = \sum_{\boldsymbol{a}\in A} \prod_{i=1}^M (y_i)_{a_i}
\end{align}
The negative log-likelihood of $p(\boldsymbol{t}|\boldsymbol{x})$ is the CTC loss function $\ell_{\text{CTC}}(f(\boldsymbol{x}), \boldsymbol{t})$. 
\begin{align}
    \ell_{\text{CTC}}(f(\boldsymbol{x}), \boldsymbol{t}) = -\log p(\boldsymbol{t}|\boldsymbol{x})
    \label{con:ctc_loss} 
\end{align}

To obtain the most probable output sequence $\mathop{\arg\max}_{\boldsymbol{t}} p(\boldsymbol{t}|\boldsymbol{x})$, a greedy path decoding algorithm can select the most probable alignment at each step. However, the greedy algorithm does not guarantee to find the most probable labeling. A better method, beam search decoding, simultaneously keeps a certain number of the most probable alignments at each step and chooses the most probable output in the top-alignment list.

\subsection{Attacking DNN-based computer vision tasks}

\subsubsection{Where to add perturbations?  }
Attacking DNN models is a popular topic in both computer vision and security fields.  
Many projects focus on finding small $L_p$-bounded perturbations, hoping that the bound $L_p$ will keep the perturbations visually imperceptible. FGSM~\cite{goodfellow2014explaining}, L-BFGS~\cite{szegedy2013intriguing}, DeepFool~\cite{moosavi2016deepfool}, Carlini~{$L_2, L_\infty$}~\cite{carlini2017towards}, PGD~\cite{madry2017towards} and EAD~\cite{chen2018ead} all performed modifications at the pixel level by a small amount bounded by $\epsilon$.  

Other attacks such as JSMA~\cite{papernot2016limitations}, Carlini $L_0$~\cite{carlini2017towards}, Adversarial Patch~\cite{brown2017adversarial} and LaVAN~\cite{karmon2018lavan}, perturb a small region of pixels in an image but the pixel-level perturbations are not bounded by $L_p$. 

As we have mentioned, neither approach can hide perturbations from the normal human vision in OCR tasks, as a document with enough readability usually has a spotless background and vivid text, which is greatly different from natural RGB photos.

\subsubsection{How to generate perturbations?  } There are two types of methods to generate perturbations.  

\para{1) Gradient-based attack} is to add perturbations generated by gradient against input pixels. 
Formally, we can describe the general problem as:
For a $L_\infty$-bounded adversary, we compute an adversarial example $\boldsymbol{x}'$ given an original image $\boldsymbol{x}$ and target labels $t$ where perturbations' bound $\epsilon > 0$ is tiny enough to be indistinguishable to human observers.

\emph{Fast Gradient Sign Method} (FGSM)~\cite{goodfellow2014explaining} is a one-step attack that obtains the adversarial image $\boldsymbol{x}'$ as $\boldsymbol{x} + \epsilon \cdot \text{sign}(\nabla_{\boldsymbol{x}} \ell(\boldsymbol{x}', t))$. The original image $\boldsymbol{x}$ takes a gradient sign step with step size $\epsilon$ in the direction that increases the probability of the target label $t$. It is efficient, but it only provides a coarse approximation of the optimal perturbations. 

\emph{Basic Iterative Method} (BIM)~\cite{kurakin2016adversarial} takes multiple smaller steps $\alpha$ and the result image is clipped by the same bound $\epsilon$: $\boldsymbol{x}'_i = \boldsymbol{x}'_{i-1}  + \text{clip}_{\epsilon}(\alpha \cdot \text{sign}(\nabla_{\boldsymbol{x}} \ell(\boldsymbol{x}'_{i-1}, t)))$, where $\boldsymbol{x}'_i$ is an adversarial example yielded at step $i$. BIM produces superior results to FGSM. 

\emph{Momentum Iterative Method} (MIM)~\cite{dong2018boosting} extends BIM with a momentum item. MIM can not only stabilize update directions but also escape from poor local maxima during the iteration. Thus, it generates more transferable adversarial examples. Each iterative update is to adjust the update direction and generate new adversarial image $\boldsymbol{x}'_i$ using the momentum item $\boldsymbol{g}_i$, as following $\boldsymbol{g}_i = \mu \cdot \boldsymbol{g}_{i-1} + \frac{\nabla_{\boldsymbol{x}} \ell(\boldsymbol{x}'_{i-1}, t)}{\|\nabla_{\boldsymbol{x}} \ell (\boldsymbol{x}'_{i-1}, t)\|_p}, \boldsymbol{x}'_i = \boldsymbol{x}'_{i-1} + \text{clip}_{\epsilon}(\alpha \cdot \text{sign}(\boldsymbol{g}_{i}))$, where $\mu$ is the decay factor.

\para{2) Optimization-based attack } directly solves the optimization problem of minimizing the $L_p$ distance between the original example $\boldsymbol{x}$ and the adversarial example $\boldsymbol{x}'$ and yielding the incorrect classification.  
\begin{align*}
    \begin{array}{ll}
    {\text { minimize }} & {\|\boldsymbol{x}'-\boldsymbol{x}\|_{p}} \\ 
    {\text { subject to }} & {f(\boldsymbol{x}')=t} \text { and }
     {\boldsymbol{x}' \in[\boldsymbol{x}_{\min},\boldsymbol{x}_{\max}]^{|\boldsymbol{x}|}}
    \end{array}.
\end{align*}

\emph{Box-constraint L-BFGS}~\cite{szegedy2013intriguing} finds the adversarial examples by solving the box-constraint problem, $\text{minimize} ~\lambda \left\|\boldsymbol{x}'-\boldsymbol{x}\right\|_{p}+\ell(\boldsymbol{x}', t)$ subject to $\boldsymbol{x}' \in [\boldsymbol{x}_{\min}, \boldsymbol{x}_{\max}]^{|\boldsymbol{x}|}$, where $\ell(\boldsymbol{x}', t)$ is the cross-entropy loss between the logit output and the target label $t$. Although the perturbations generated by L-BFGS are much less than the gradient-based attack, L-BFGS has far low efficiency. 

\emph{C\&W}~\cite{carlini2017towards} is a $L_p$-oriented attack that can successfully break undefended and defensively distilled DNNs. Given the logit $Z$ of the model $f$, other than applying cross-entropy as the loss function, C\&W attack designed a new loss function $\ell(\boldsymbol{x}', t) = \max(\max\{Z(\boldsymbol{x}')_i: i \neq t\} - Z(\boldsymbol{x}')_t, -\kappa)$ solved by gradient descent, where $\kappa$ is the confidence of misclassification. 

\subsection{Defense methods against these attacks}

People have proposed many practical defense methods against adversarial examples.  
\emph{Adversarial training}~\cite{tramer2017ensemble} improves the robustness of DNNs by injecting label-corrected adversarial examples into the training procedure and training a robust model that has a resistance to perturbations generated by gradient-based methods.  
\emph{Defensive distillation}~\cite{papernot2016distillation} defends against adversarial perturbations using the distillation techniques~\cite{hinton2015distilling} to retrain the same network with class probabilities predicted by the original network. 
There are also methods focusing on detecting adversarial samples~\cite{Xu2017Feature,lu2017safetynet,grosse2017statistical,feinman2017detecting}. 


\section{Methodology}

\para{Preliminaries.} We assume that attackers have full knowledge of the threat model, such as the model architecture and parameter values. Given an input image $\boldsymbol{x} \in [\boldsymbol{x}_{\min}, \boldsymbol{x}_{\max}]^{|\boldsymbol{x}|}$, an adversarial image $\boldsymbol{x}' \in [\boldsymbol{x}_{\min}, \boldsymbol{x}_{\max}]^{|\boldsymbol{x}|}$ where $|\boldsymbol{x}|$ is the number of pixels in the original image, OCR model $f$, the adversarial example's prediction result is $f(\boldsymbol{x}')$. Given target label $\boldsymbol{t}$, loss function of the target model $f$ with respect to the input image $\boldsymbol{x}$ is $\ell(\boldsymbol{x}, \boldsymbol{t})$. Besides, we assume $f(\boldsymbol{x})$ to be differentiable almost everywhere (sub-gradient may be used at discontinuities). Because the gradient-descent approach is applicable to any DNNs with a differentiable discriminant function.

\para{Distance Metric.} We define the distance metric $L_p$ to quantify the similarity between the original image $\boldsymbol{x}$ and the corresponding adversarial image $\boldsymbol{x}'$. Such a distance metric reflects the cost of manipulations. $L_p$-norm is a widely-used distance metric defined as $\| {x} - {x}' \|_p = \sqrt[p]{\sum_{i=1}^{d} |\boldsymbol{x}_i-\boldsymbol{x}'_i|^p}$ where d-dimensional vector $\boldsymbol{x} = [x_1, ..., x_d]$ for any $p \in \{0, 1, 2, \infty \}$. $L_1$ accounts for the total variation in the perturbations and serves as a popular convex surrogate function of the $L_0$ that measures the number of modified pixels. $L_2$ measures the standard Euclidean distance, which is usually used to improve the visual quality. $L_\infty$ measures the maximum change of the perturbations between $\boldsymbol{x}$ and $\boldsymbol{x}'$.

    

\begin{figure*}[]
\centering
\includegraphics[width=0.8\textwidth]{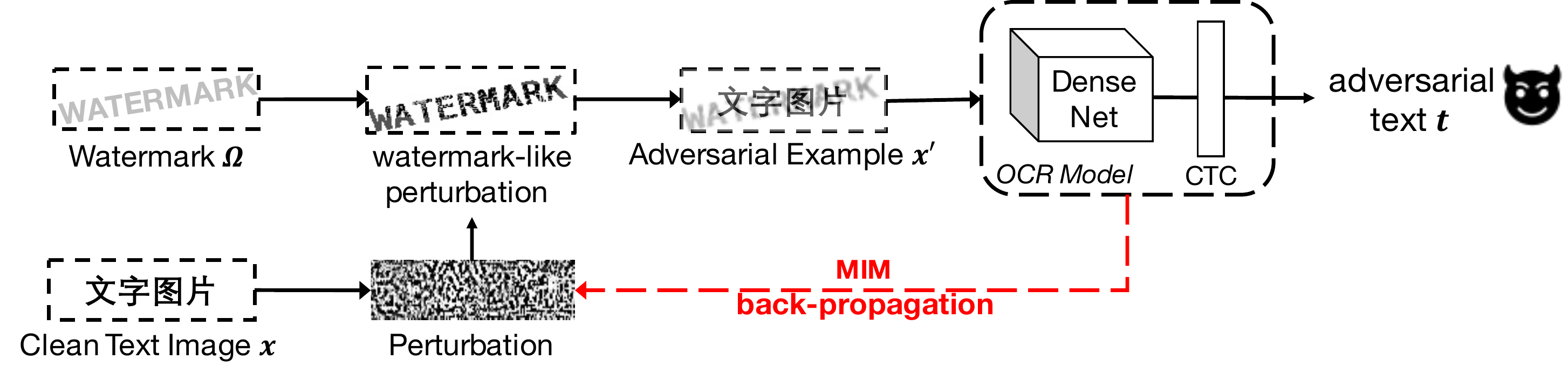}
\caption{\small The pipeline of the {\wm} attack. We generate noise using MIM with CTC loss function back propagating the targeted DenseNet and then mask the noise outside the watermark region. We iterating the procedure above until an iterations threshold.}
\label{fig:flowchart}
\end{figure*}

\subsection{{\wm} attack to CTC-based OCR}

In this paper, we propose the MIM-based {\wm} attack on the CTC-based OCR model to generate adversarial examples. In this section, we will first introduce how to integrate watermarks into MIM~\cite{dong2018boosting} , which induces the MIM-based {\wm} attack method (WM) to generate adversarial examples satisfying the $L_{\infty}$-norm restriction in the targeted and non-targeted attack fashion. We then present several variants of WM to $L_{\infty}$-norm bound. The generation pipeline of the {\wm} adversarial attack is illustrated in Figure \ref{fig:flowchart}. Table \ref{tab:adv_example} shows adversarial examples generated from each method.
 
 \subsubsection{MIM-based {\wm} attack (WM).}
 Watermark widely occurs in a mass of documents and files. Making use of the popularity of the watermark in the documents, we apply the idea of the watermark to decorate perturbations as the watermark; that is, we restrict the manipulation region on a specific predefined watermark-shape region $\Omega$.  

To generate a targeted $L_{\infty}$-bounded adversarial example $\boldsymbol{x}'$, we start with an initial image $\boldsymbol{x}'_0=\boldsymbol{x}$ given an original image $\boldsymbol{x}$. WM seeks the adversarial example by solving the constrained optimization problem
 \begin{align*}
     \underset{\boldsymbol{x}'}{\arg \min }~&\ell_{\text{CTC}}(\boldsymbol{x}', \boldsymbol{t}) \\
     \text { s.t. } & \|\boldsymbol{x}'-\boldsymbol{x}\|_{\infty} \leq \epsilon \text{ and }
     (1-\Omega) \odot (\boldsymbol{x}'-\boldsymbol{x}) = \boldsymbol{0},
 \end{align*}
where $\epsilon$ is the size of adversarial perturbations and $\Omega = \{o\in \Omega | o=1 \text{ inside the watermark, otherwise, } o=0, \Omega\in \{0, 1\}^{|\boldsymbol{x}|}\}$. 
We summarized WM in Algorithm \ref{alg:1}.
At each attacking iteration $i$, the attacker first feeds the adversarial example $\boldsymbol{x}'_i$ to the OCR model $f$ and obtain the gradient $\nabla_{\boldsymbol{x}} \ell_{\text{CTC}} (\boldsymbol{x}_i', \boldsymbol{t})$ through back-propagation. Then, for the purpose of stabilizing update directions and escaping from poor local maxima, update momentum item $\boldsymbol{g}_{i+1}$ by accumulating the velocity vector in the gradient direction as Equation \ref{eq:momentum} shown in Algorithm \ref{alg:1}. Last, update new adversarial example $\boldsymbol{x}_{i+1}'$ by applying the $\Omega$-restricted sign gradient with small step size $\alpha$, and clip the intermediate perturbations to ensure them in the $\epsilon$-ball of $\boldsymbol{x}'_0$ as Equation \ref{fmt:adv_x}. The attacking iteration proceeds until the attack is successful or reaches the maximum iterations $I$.

\begin{algorithm}[tb]
    \footnotesize
    \renewcommand{\algorithmicrequire}{\textbf{Input:}}
    \renewcommand{\algorithmicensure}{\textbf{Output:}}
    \caption{MIM-based {\wm} example generation}
    \label{alg:1}
    \begin{algorithmic}[1]
        \REQUIRE A clean image $\boldsymbol{x}$, OCR model $f$ with CTC loss $\ell_{\text{CTC}}$, ground-truth text $f(\boldsymbol{x})$, targeted text $\boldsymbol{t}$, watermark modification region $\Omega$, $\epsilon$-ball perturbation, \# of iterations $I$, decay factor $\mu$
        \ENSURE An adversarial example $\boldsymbol{x}'$ with $\|\boldsymbol{x}' - \boldsymbol{x}\|_{\infty} \leq \epsilon$ or attack failure $\bot$ 
        \STATE Initialization: $\alpha = \epsilon / I$; $\boldsymbol{g}_0 = \boldsymbol{0}$; $\boldsymbol{x}_0' = \boldsymbol{x}$
        \FORALL{each iteration $i = 0$ to $I-1$}
        \STATE Input $\boldsymbol{x}'_{i}$ to $f$ and obtain the gradient $\nabla_{\boldsymbol{x}} \ell_{\text{CTC}}(\boldsymbol{x}'_i, \boldsymbol{t})$
        \STATE Update $\boldsymbol{g}_{i+1}$ by accumulating the velocity vector in the gradient direction as 
        \begin{align} 
        \label{eq:momentum}
        \boldsymbol{g}_{i+1}=\mu \cdot \boldsymbol{g}_{i}+\frac{\nabla_{\boldsymbol{x}} \ell_{\text{CTC}}(\boldsymbol{x}_{i}', \boldsymbol{t})}{\|\nabla_{\boldsymbol{x}} \ell_{\text{CTC}}(\boldsymbol{x}_{i}', \boldsymbol{t})\|_{p}} 
        \end{align}
        \STATE Update $\boldsymbol{x}_{i+1}'$ by applying watermark-bound sign gradient as 
        \begin{align} 
        \label{fmt:adv_x}
        \boldsymbol{x}_{i+1}' = \boldsymbol{x}_{i}' + \text{clip}_{\epsilon}(\alpha \cdot \Omega \odot \operatorname{sign}\left(\boldsymbol{g}_{i+1}\right))
        \end{align}
        \IF{$f(\boldsymbol{x}'_{i+1}) == t$}
        \RETURN $\boldsymbol{x}' = \boldsymbol{x}'_{i+1}$
        \ENDIF
        \ENDFOR
        \RETURN failure $\bot$
\end{algorithmic}
\end{algorithm}

\begin{table*}[t]
\begin{center}

\begin{tabular}{| c | c | c | c | c | c | c | c | c |} 
\hline
 & original & MIM & WM & {WM\tiny{init}} & {WM\tiny{neg}} & {WM\tiny{edge}} & OCR output & English translation\\ \hline

  \multirow{5}{*}{\rotatebox{90}{ substitution \qquad \quad }} &
 \begin{minipage}[]{0.07\textwidth}
 \centerline{ \includegraphics[width=\linewidth]{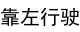} }
  \end{minipage} & 
  \begin{minipage}[]{0.07\textwidth}
   \centerline{\includegraphics[width=\linewidth]{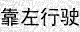}}
    \end{minipage}  & 
   \begin{minipage}[]{0.1\textwidth}
 \centerline{ \includegraphics[width=\linewidth]{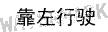}}
  \end{minipage} & 
  \begin{minipage}[]{0.1\textwidth}
 \centerline{ \includegraphics[width=\linewidth]{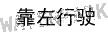}}
  \end{minipage} & 
  \begin{minipage}[]{0.1\textwidth}
 \centerline{ \includegraphics[width=\linewidth]{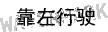}}
  \end{minipage} & 
  \begin{minipage}[]{0.07\textwidth}
 \centerline{ \includegraphics[width=\linewidth]{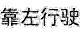}}
  \end{minipage} &
  \begin{minipage}[]{0.09\textwidth}
 \centerline{ \includegraphics[width=\linewidth]{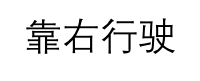}}
  \end{minipage}  &
  \tabincell{c}{\small{drive left}$\rightarrow$\\ \small{drive right}}
  \\  
  
 &
  \begin{minipage}[]{0.07\textwidth}
 \centerline{ \includegraphics[width=\linewidth]{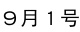}}
  \end{minipage} & 
  \begin{minipage}[]{0.07\textwidth}
   \centerline{\includegraphics[width=\linewidth]{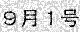}}
    \end{minipage}  & 
   \begin{minipage}[]{0.1\textwidth}
 \centerline{ \includegraphics[width=\linewidth]{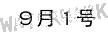}}
  \end{minipage} & 
  \begin{minipage}[]{0.1\textwidth}
 \centerline{ \includegraphics[width=\linewidth]{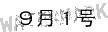}}
  \end{minipage} & 
  \begin{minipage}[]{0.1\textwidth}
 \centerline{ \includegraphics[width=\linewidth]{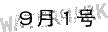}}
  \end{minipage} & 
  \begin{minipage}[]{0.07\textwidth}
 \centerline{ \includegraphics[width=\linewidth]{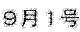}}
  \end{minipage} &
  \begin{minipage}[]{0.09\textwidth}
 \centerline{ \includegraphics[width=\linewidth]{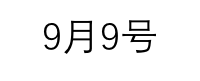}}
  \end{minipage}  &
   \tabincell{c}{\small{1 Sep.}$\rightarrow$\small{9 Sep.}}
   \\  
  
  &
  \begin{minipage}[]{0.07\textwidth}
 \centerline{ \includegraphics[width=\linewidth]{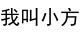}}
  \end{minipage} & 
  \begin{minipage}[]{0.07\textwidth}
   \centerline{\includegraphics[width=\linewidth]{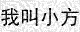}}
    \end{minipage}  & 
   \begin{minipage}[]{0.1\textwidth}
 \centerline{ \includegraphics[width=\linewidth]{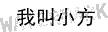}}
  \end{minipage} & 
  \begin{minipage}[]{0.1\textwidth}
 \centerline{ \includegraphics[width=\linewidth]{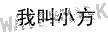}}
  \end{minipage} & 
  \begin{minipage}[]{0.1\textwidth}
 \centerline{ \includegraphics[width=\linewidth]{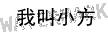}}
  \end{minipage} & 
  \begin{minipage}[]{0.07\textwidth}
 \centerline{ \includegraphics[width=\linewidth]{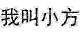}}
  \end{minipage} &
  \begin{minipage}[]{0.09\textwidth}
 \centerline{ \includegraphics[width=\linewidth]{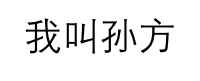}}
  \end{minipage}  &
   \tabincell{c}{\small{I am Xiao Fang}$\rightarrow$\\ \small{I am Sun Fang}}
  \\  
  
  &
\begin{minipage}[]{0.07\textwidth}
 \centerline{ \includegraphics[width=\linewidth]{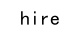}}
  \end{minipage} & 
  \begin{minipage}[]{0.07\textwidth}
   \centerline{\includegraphics[width=\linewidth]{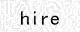}}
    \end{minipage}  & 
   \begin{minipage}[]{0.1\textwidth}
 \centerline{ \includegraphics[width=\linewidth]{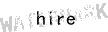}}
  \end{minipage} & 
  \begin{minipage}[]{0.1\textwidth}
 \centerline{ \includegraphics[width=\linewidth]{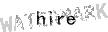}}
  \end{minipage} & 
  \begin{minipage}[]{0.1\textwidth}
 \centerline{ \includegraphics[width=\linewidth]{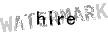}}
  \end{minipage} & 
  \begin{minipage}[]{0.07\textwidth}
 \centerline{ \includegraphics[width=\linewidth]{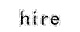}}
  \end{minipage} & 
  \begin{minipage}[]{0.09\textwidth}
 \centerline{ \includegraphics[width=\linewidth]{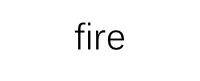}}
  \end{minipage}   &
\small{hire}$\rightarrow$\small{fire}
  \\  
  
    &
\begin{minipage}[]{0.07\textwidth}
 \centerline{ \includegraphics[width=\linewidth]{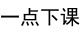}}
  \end{minipage} & 
  \begin{minipage}[]{0.07\textwidth}
   \centerline{\includegraphics[width=\linewidth]{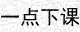}}
    \end{minipage}  & 
   \begin{minipage}[]{0.1\textwidth}
 \centerline{ \includegraphics[width=\linewidth]{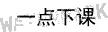}}
  \end{minipage} & 
  \begin{minipage}[]{0.1\textwidth}
 \centerline{ \includegraphics[width=\linewidth]{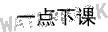}}
  \end{minipage} & 
  \begin{minipage}[]{0.1\textwidth}
 \centerline{ \includegraphics[width=\linewidth]{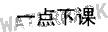}}
  \end{minipage} & 
  \begin{minipage}[]{0.07\textwidth}
 \centerline{ \includegraphics[width=\linewidth]{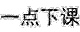}}
  \end{minipage} & 
  \begin{minipage}[]{0.09\textwidth}
 \centerline{ \includegraphics[width=\linewidth]{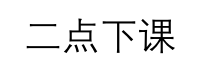}}
  \end{minipage} & 
   \tabincell{c}{\small{class is over at 1}$\rightarrow$\\ \small{class is over at 2}}
  \\  \hline
  
$-$  &
\begin{minipage}[]{0.07\textwidth}
 \centerline{ \includegraphics[width=\linewidth]{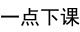}}
  \end{minipage} & 
  \begin{minipage}[]{0.07\textwidth}
   \centerline{\includegraphics[width=\linewidth]{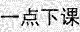}}
    \end{minipage}  & 
   \begin{minipage}[]{0.1\textwidth}
 \centerline{ \includegraphics[width=\linewidth]{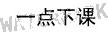}}
  \end{minipage} & 
  \begin{minipage}[]{0.1\textwidth}
 \centerline{ \includegraphics[width=\linewidth]{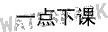}}
  \end{minipage} & 
  \begin{minipage}[]{0.1\textwidth}
 \centerline{ \includegraphics[width=\linewidth]{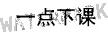}}
  \end{minipage} & 
  \begin{minipage}[]{0.07\textwidth}
 \centerline{ \includegraphics[width=\linewidth]{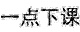}}
  \end{minipage} & 
  \begin{minipage}[]{0.09\textwidth}
 \centerline{ \includegraphics[width=\linewidth]{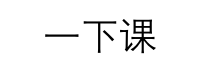}}
  \end{minipage} &
   \tabincell{c}{\small{class is over at 1}$\rightarrow$\\ \small{once class is over}}
  \\  \hline

$+$ &
\begin{minipage}[]{0.07\textwidth}
 \centerline{ \includegraphics[width=\linewidth]{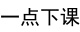}}
  \end{minipage} & 
  \begin{minipage}[]{0.07\textwidth}
   \centerline{\includegraphics[width=\linewidth]{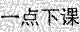}}
    \end{minipage}  & 
   \begin{minipage}[]{0.1\textwidth}
 \centerline{ \includegraphics[width=\linewidth]{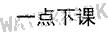}}
  \end{minipage} & 
  \begin{minipage}[]{0.1\textwidth}
 \centerline{ \includegraphics[width=\linewidth]{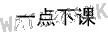}}
  \end{minipage} & 
  \begin{minipage}[]{0.1\textwidth}
 \centerline{ \includegraphics[width=\linewidth]{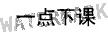}}
  \end{minipage} & 
  \begin{minipage}[]{0.07\textwidth}
 \centerline{ \includegraphics[width=\linewidth]{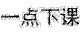}}
  \end{minipage} & 
  \begin{minipage}[]{0.09\textwidth}
 \centerline{ \includegraphics[width=\linewidth]{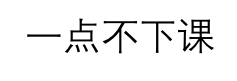}}
  \end{minipage} &
   \tabincell{c}{\small{class is over at 1}$\rightarrow$\\ \small{class is not over at 1}}
  \\  \hline

\end{tabular}
\end{center}
\caption{\small Adversarial examples with different attacks.  The last two rows show text deletion and insertion.  Other rows show text substitution.}
\label{tab:adv_example}
\end{table*}

\subsubsection{Variants of WM attack.}

To present a more natural appearance of watermark-like perturbations, we design three variants of WM attack.

\para{\textbf{WM\tiny{init}}.} Watermark region $\Omega$ element-wisely multiplies the sign gradient $\operatorname{sign} (\boldsymbol{g}_{i+1})$ which attackers only operate pixels inside the watermark region $\Omega$. As shown in the WM column of Table \ref{tab:adv_example}, the perturbations of WM adversarial examples are not dense enough to construct a complete watermark and be a natural watermark. Thus, for filling in the blanks of the watermark region, we start from an initial watermark-pasted image by attaching a watermark to the original image, $\boldsymbol{x}_0' = \boldsymbol{x} + \lambda \cdot (\boldsymbol{x} > \tau) \odot \Omega$, where $\lambda$ is the grayscale value of the pasted watermark and $(\boldsymbol{x} > \tau)$ denotes the position except the text.

\para{\textbf{WM\tiny{neg}}.} The sign of the gradient, $\operatorname{sign}(\cdot )$, can be -1 or +1 based on the direction of gradient descent. When the gradient is positive, $\operatorname{sign}(\cdot )=1$, the pixel value will increase, that is, the pixel looks whiter (the maximum of the grayscale value mean the whitest color or else blackest color). Otherwise, the pixel value decreases, and the pixel looks blacker. Obviously, the pixels in the text region become whiter resulting in the fuzzy text, and it's meaningless to whiten the clear white background. Thus, we only need to keep the negative gradient and leave the positive gradient behind. We generated WM noise but only kept the negative gradient during attacking iteration. After adding the new constraint, the update step of new adversarial example, Equation \ref{fmt:adv_x}, becomes 
\begin{align}
    \boldsymbol{x}_{i+1}' = \boldsymbol{x}_{i}' + \text{clip}_{\epsilon}(\alpha \cdot \Omega \odot \min(0, \operatorname{sign}\left(\boldsymbol{g}_{i+1}\right))).
\end{align}


\para{\textbf{WM\tiny{edge}}.} A different way to add perturbations is to confine watermark region around the text edges, pretending to be defectives in printing.

{WM\tiny{edge}} is similar with WM. We define the watermark as the region of the text edge, which can be obtained by image erosion in morphological image processing. The erosion operation $\ominus$
erodes the original image using the specified structuring element that determines the shape of a pixel neighborhood over which the minimum is taken. In experiments, we use a  $3 \times 3$ rectangular structuring element as a kernel, $\mathcal{K} := \boldsymbol{1}^{3\times3}$. We take the bolder text region after erosion as the watermark. Thus, the text-edge shape watermark $\Omega_{\text{edge}}$ is defined as $\Omega_{\text{edge}} = \{o\in\Omega_{\text{edge}} | o=\begin{cases}
1, & \mbox{if } \boldsymbol{x} \ominus \mathcal{K} > \tau \\
0, & \mbox{if } \boldsymbol{x} \ominus \mathcal{K} \leq \tau
\end{cases}
\}.$
%




\section{Experiments}

In this section, we generate adversarial examples on the CTC-based OCR model. We compared the performance of the basic MIM, WM, and its variants. 

\subsection{Setup}

\para{Threat model.} We performed WM attack on the DenseNet + CTC neural network~\footnote {\url{https://github.com/YCG09/chinese_ocr}} which is trained in the Chinese text image dataset. DenseNet \cite{huang2017densely} is one of powerful DNNs for visual recognition, which can capture complex features. Thus, we utilize DenseNet as the feature extractor and CTC \cite{graves2006connectionist} as the loss function. In the test phase, the DenseNet+CTC OCR recognition model achieved 98.3\% accuracy on the validation dataset that involves 36400 images. The Chinese text image dataset has 3.64 million images that are generated by altering fonts, scale, grayscale, blur, sketch based on Chinese news articles. The character set has 5989 unique characters, including Chinese and English characters, digits, punctuations.

\para{Attack setting.} The attack setting is applied among all experiments. Our experiment setup is based on MIM's framework. We use the implementation of MIM in CleverHans package\footnote{\url{https://github.com/tensorflow/cleverhan}}. We use the attack setting which runs $I=1000$ iterations at the most. We utilize an early stopping criterion based on the attacking result at each iteration. The $L_{\infty}$-norm perturbations $\epsilon$ is bounded by 0.2. The pixel value of the image ranges from 0 (black) to 1 (white). For the initial watermark in {WM\tiny{init}}, we set the grayscale value $\lambda$ of the watermark to 0.3, and we put the watermark in the center of the image by default. The watermarks are set to the font size 30.

\para{Choose attacked candidates.} To satisfy the semantic fluency in our OCR attack, we choose 691 pairs of antonym characters with high similarity of character shape, $\text{Sim}(c, c') > 0.6$ \footnote{Given two characters $c$ and $c'$, character similarity is defined as $\text{Sim}(c, c') = w_1 \text{Stroke}(c, c') + w_2 \text{Sijiao}(c, c') + w_3 \text{Edit}(c, c') $ \\ where $\text{Stroke}(c, c')$ denotes the absolute value of strokes between $c$ and $c'$. $\text{Sijiao}(c, c')$ is the Levenshtein distance of sijiao between $c$ and $c'$, which is an encoding approach to fast retrieving Chinese characters. $\text{Edit}(c, c')$ is edit distance of character images between $c$ and $c'$. The weights $w_1$, $w_2$ and $w_3$ is chosen as 0.33, 0.33 and 0.34, respectively.}, so that the adversarial attack only requires adversarial perturbations as less as possible to fool the OCR system. Then we match selected antonym character pairs in the corpus of People's Daily in1988 and choose 158 sentences containing selected characters which do not cause syntactic errors after substituting the corresponding antonym character. Last, we generate the line image containing chosen sentences. 

\para{Evaluation metrics.} To quantify perturbations of adversarial images $\boldsymbol{x}'$ compared with benign images $\boldsymbol{x}$, we measure perturbations from MSE, PSNR and SSIM. 
\textit{Mean-squared error} (MSE) denotes the difference between adversarial images  and original images, calculated by 
$\mathrm{MSE} = \frac{1}{|\boldsymbol{x}|} \|\boldsymbol{x} - \boldsymbol{x}'\|_2$.
 \textit{Peak-signal-to-noise ratio} (PSNR) is a ratio of maximum possible power of a signal and power of distortion, calculated by $\operatorname{PSNR}=10 \log \left(\frac{D^{2}}{\mathrm{MSE}}\right)$ where $D$ denotes the dynamic range of pixel intensities, e.g., for an 8 bits/pixel image we have $D = 255$.
\textit{Structural similarity index} (SSIM) attempts to model the structural information of an image from luminance, contrast and structure respectively.

To evaluate the efficiency of adversarial attacks, we calculate \textit{attack success rate} (ASR) by $\mathrm{ASR}=\frac{\#(f(\boldsymbol{x}')=t) + \#(f(\boldsymbol{x})\neq f(\boldsymbol{x}'))}{\# (\boldsymbol{x})}$ that is the fraction of adversarial images that were successful in fooling the DNN model, \textit{targeted attack success rate} (ASR*) of adversarial attacks calculated as $\mathrm{ASR}^*=\frac{\#(f(\boldsymbol{x}')=t)}{\# (\boldsymbol{x})}$, the average time to generate adversarial perturbations from the clean images.

\subsection{Comparison of attacks on single character altering}

We compare different methods of altering a single character.  Table 1 shows some successful adversarial examples that different attack methods generated.  Our intuitions are:
1) MIM generates human-perceptible and unnatural noise on account of the dirty background, which distributes all over the image, and harms the image structure similarity and image quality. 
2) WM and its variants retain the noise in the watermark region bringing in a more clear background and reasonable perturbations. 
3) The watermark-fashion perturbations of WM are relatively light, and do not look like a real watermark.
4) {WM\tiny{init}} and {WM\tiny{neg}} look more real with darker and more complete watermark's shape. 
5) The perturbations of {WM\tiny{edge}} are around the edge of the text, which makes the text looks bolder and similar to printing/scanning defects. 

Intuitively, we can see that WM family of attacks generate better visual quality (in terms of looking natural) images if the attack is successful.

We evaluate the attack performance of altering a single character using the corpus discussed above. In Table \ref{tab:attack_result}, we report the metrics above (MSE/PSNR/SSIM,ASR*/ASR), as well as the average time required to generate an attack sample.  Our observations are:
1) From Table \ref{tab:attack_result} , compared to MIM, we can observe that WM, {WM\tiny{neg}} and {WM\tiny{edge}} obtained lower MSE, higher PSNR and higher SSIM, indicating that the noise level is indeed lower on a successful attack.  
2) Due to the lower noise level, the attack success rates (ASR* and ASR) of WM and its variants are also lower than MIM's.  We believe there are several reasons why they are lower.  First, in this preliminary study, we always choose a fixed shape and location of the watermark that is at the center of the original image.  The fixed location severely limits what the adversary can do.  As our future work, we will allow multiple shapes of the watermark (e.g., different texts, logos of the watermark), and different locations.  
3) {WM\tiny{neg}} casts away the positive gradient noise which possesses a certain attack ability. Hence, {WM\tiny{neg}} behaved worse than WM.  However, {WM\tiny{neg}} does generate more natural examples visually.  
4) {WM\tiny{edge}} is a special case of WM which restricts the watermark region to the shape of the text edge, which has no location problem of the watermarks like other WM-series methods. It achieved good ASR and preserved the naturalness of perturbations. It's obvious to find that retaining all gradient noise is better than only keeping the negative gradient noise. 
5) {WM\tiny{0}} is to attach an initial watermark to the original image, which can evaluate the impact of the watermark originally. After attaching an initial watermark, 37.9\% and 17.1\% images are misclassified by the DenseNet+CTC model and Tesseract OCR, respectively. Thus watermark owns attacking properties intrinsically.
6) The time for producing each adversarial example is similar and within a reasonable range.  Thus, a practical strategy is to combine different attack methods to improve ASR.

\begin{table*}[h]
\centering
\footnotesize
\begin{tabular}{|c|ccc|cc|c|cc|}
\hline
\multirow{2}{*}{} & \multicolumn{6}{c|}{\textbf{DenseNet+CTC}}  &\multicolumn{2}{c|}{\textbf{Tesseract OCR}} \\ \cline {2-9}

 & \textbf{MSE} & \textbf{PSNR} & \textbf{SSIM} & \textbf{ASR*} & \textbf{ASR} & \textbf{Time (s)} & \textbf{ASR*}       & \textbf{ASR}    \\ \hline
\textbf{MIM}                           & 0.0102 & 32.27 & 84.92 & 92.4 & 93.7 & 20.8 & 19.6 & 84.2 \\ \hline
\textbf{WM}                   & 0.0020 & 34.70 & 96.03 & 60.8 & 61.4 & 17.5 & 19.0 & 88.0\\
\textbf{WM\tiny{neg}}    & 0.0023 & 34.26 & 94.11 & 52.5 & 53.2 & 20.7 & 19.6 & 88.6\\
\textbf{WM\tiny{init}}     & 0.0094 & 34.64 & 89.02 & 55.1 & 71.5 & 19.9 &  0.0 &  100.0 \\
\textbf{WM\tiny{edge}}  & 0.0058 & 34.43 & 93.87 & 87.3 & 88.0  & 13.5 & 19.6 & 83.5 \\ 
\textbf{WM\tiny{0}}  &0.0034 & 30.92 & 93.37 &   & 37.9 &  &  & 17.1 \\ 
\hline
\end{tabular}
\caption{\small Comparison of varing adversarial attacks on DenseNet+CTC and Tesseract OCR. See text for the metric definitions. }
\label{tab:attack_result}
\end{table*}

\begin{table*}[th]
\centering
\footnotesize
\begin{tabular}{|c|c|c|cccc|c|}
\hline
\multicolumn{2}{|c|}{\textbf{ASR*/ASR}}   & \textbf{MIM } & \textbf{WM}& \textbf{WM\tiny{neg}} & \textbf{WM\tiny{init}} & \textbf{WM\tiny{edge}} & \textbf{example} \\ \hline
\multicolumn{2}{|c|}{\textbf{no deformation}} 
                                                  & 92.4/93.7 & 60.8/61.4 & 52.5/53.2 & 55.1/71.5 & 87.3/88.0 & \begin{minipage}[]{0.09\textwidth}
 \centerline{ \includegraphics[width=0.8\linewidth]{./right_side-original.jpg}}
  \end{minipage} \\ \hline
\multirow{6}{*}{\rotatebox{90}{ \textbf{Deformation}}} 
& \textbf{AvgBlur@2x2}               & 44.30/62.7   & 47.47/51.27 & 46.84/54.43 & 18.99/79.11   & 48.73/55.06  & \begin{minipage}[]{0.09\textwidth}
 \centerline{ \includegraphics[width=\linewidth]{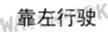}}
  \end{minipage} \\
& \textbf{MedianBlur@3x3}             & 2.53/99.37    & 1.27/98.73  & 1.90/95.57  & 0.00/100.0  & 2.53/96.84  & \begin{minipage}[]{0.09\textwidth}
 \centerline{ \includegraphics[width=\linewidth]{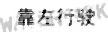}}
  \end{minipage} \\
& \textbf{GaussianBlur@3x3,0}   & 51.27/58.86 & 48.10/52.53 & 46.20/55.70 & 27.85/72.78 & 52.53/55.06  & \begin{minipage}[]{0.09\textwidth}
 \centerline{ \includegraphics[width=\linewidth]{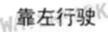}}
  \end{minipage} \\
& \textbf{Salt\&Pepper@2\%}       & 4.43/99.37   & 3.16/98.10  & 2.53/97.47  & 0.00/100.0  & 3.80/98.73  & \begin{minipage}[]{0.09\textwidth}
 \centerline{ \includegraphics[width=\linewidth]{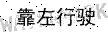}}
  \end{minipage} \\
& \textbf{Compress@20}             & 38.61/68.99 & 46.20/55.06 & 47.47/58.86 & 22.78/76.58   & 56.33/63.92 & \begin{minipage}[]{0.09\textwidth}
 \centerline{ \includegraphics[width=\linewidth]{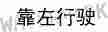}}
  \end{minipage} \\ 
& \textbf{inpainting@2}     &                      &  0.00/100.0 & 0.00/100.0 & 0.00/100.0  & 0.00/100.0 & \begin{minipage}[]{0.09\textwidth}
 \centerline{ \includegraphics[width=\linewidth]{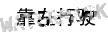}}
  \end{minipage} \\ \hline
\end{tabular}
\caption{\small Comparison of ASR*/ASR (\%) of adversarial examples under different preprocessing / defense methonds. }
\label{tab:deformation_result}
\end{table*}


\subsection{Attack transferability to blackbox OCR}

We want to see if adversarial examples can mislead other (black-box) models, or have commonly called \textit{transferability} \cite{liu2016delving,papernot2016transferability,sharma2017attacking,papernot2017practical}. 


We adopt the widely-used latest version Tesseract OCR~\cite{Tesseract} as a black-box model to perform adversarial attacks. 
We fed the adversarial samples, which are generated by attack methods above in the Densenet+CTC model, into the off-the-shelf Tesseract OCR, and evaluated recognition results~(ASR*/ASR) shown in the last two columns of  Table~\ref{tab:attack_result}.  

We find that all attacks produce transferable adversarial examples in terms of ASR.  It may be due to the reason that the noise indeed perturbs the intrinsic features of a character sequence for different models, or because Tesseract OCR cannot handle noise.  However, ASR* reduces significantly because perturbations are still trained on a different model.

\subsection{Real-world examples}

\begin{figure}[tb]
    \centering
    \includegraphics[width=0.25\textwidth]{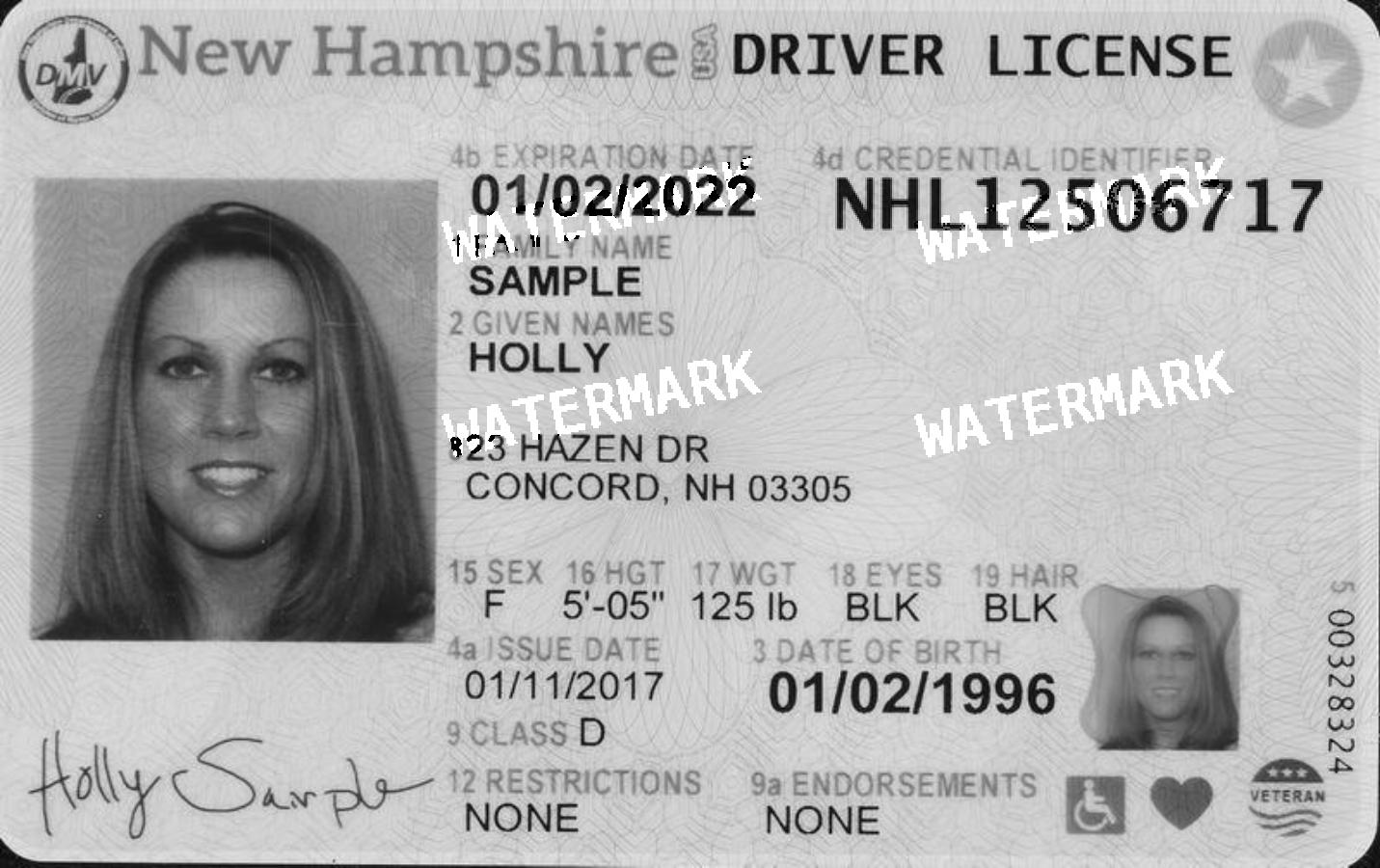}
    \caption{An adversarial attack example in driver license recognition. The OCR output a licenses number of  \textit{N\textbf{A}L1250\textbf{5}717} while it is actually \textit{N\textbf{H}L1250\textbf{6}717}.}
    \label{fig:license_sample}
\end{figure}

\begin{figure}[tb]
    \centering
    \includegraphics[width=0.505\textwidth]{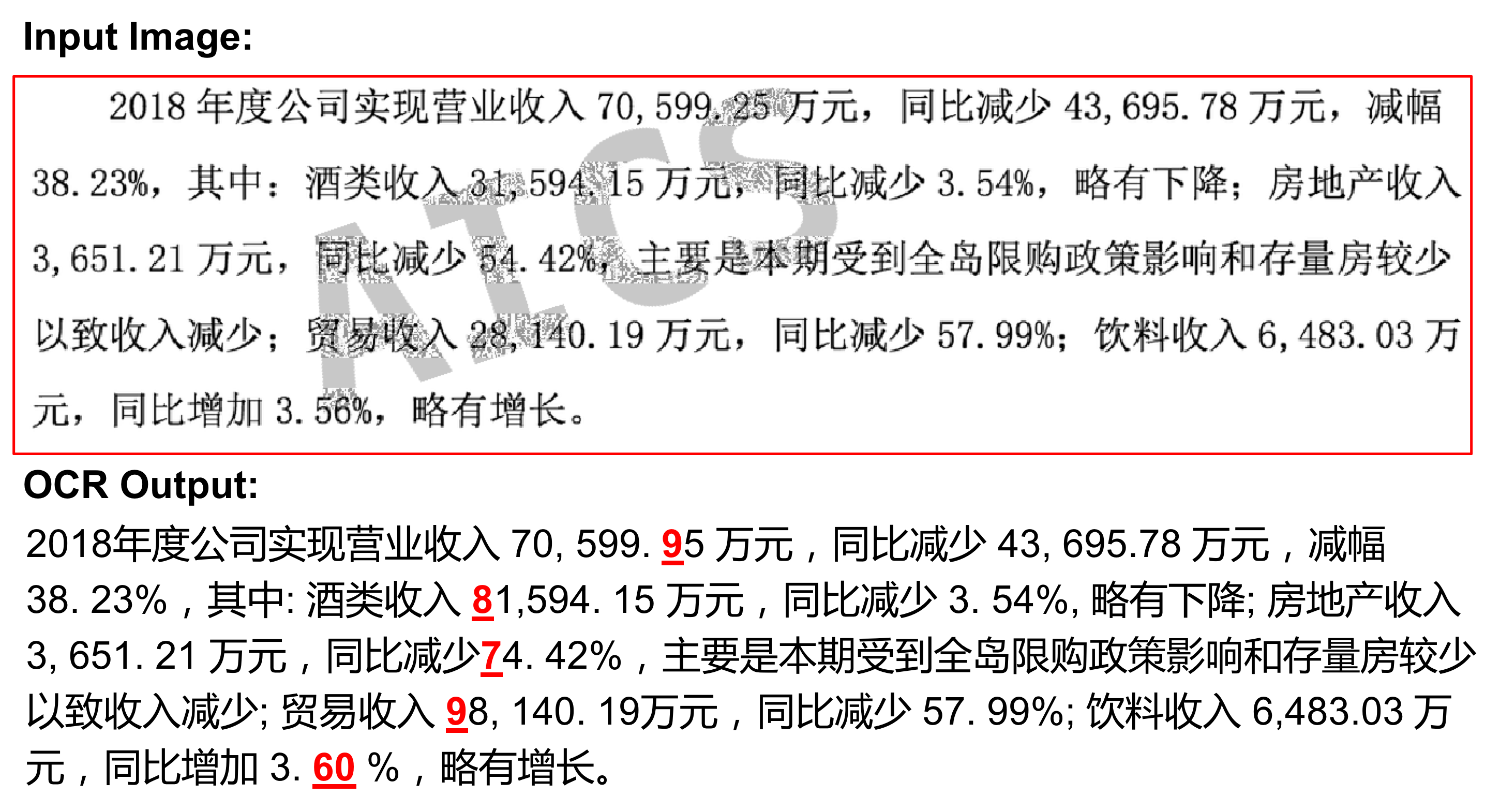}
    \caption{Attack on a listed Chinese company's annual report.  All the revenue numbers are altered in the OCR result.}
    \label{fig:report_sample}
\end{figure}

Figure~\ref{fig:license_sample} and~\ref{fig:report_sample} show two real-world examples of WM. In Figure 2, using watermarks, we successfully altered the license number recognition results.  Figure 3 shows an example of a paragraph of an annual financial report of a public company. By adding the AICS watermark, we altered all the revenue numbers in the recognition results.  

\subsection{Defense against these attacks}

We evaluate the robustness of these attacks against common defense methods that preprocess the input images, and Table~\ref{tab:deformation_result} summarizes the results.  

\para{Noise removing methods with local smoothing.}  Local smoothing makes use of nearby pixels to smooth each pixel. We use three  local smoothing methods from OpenCV \cite{OpenCVweb}, \emph{average blur}, \emph{median blur} and \emph{gaussian blur}.  We observe that
1) median blur is particularly effective in removing sparsely-occurring black and white pixels in an image while preserving edges of the text well.  
2) Median blur with kernel size 3 blurs texts so much that OCR algorithms no longer work, although it reduces ASR*.
3) Average smoothing with kernel size 2 and Gaussian smoothing with kernel size 3 have similar performance. Although MIM has a high ASR, it seems more sensitive to various deformations than WM and {WM\tiny{edge}}. 

\para{Salt\&pepper noise } is a common way against adversarial examples.  We find that it is particularly effective in decreasing ASR* (to 0), but the result is an increase of ASR to almost $100\%$.  It indicates that salt\&pepper noise harms the general image quality too much in exchange for reducing adversarial perturbations.  

\para{Image compression.} We show that adversarial examples can survive lossy image compression and decompression, applying the standard JPEG  compression method, with the quality parameter is set to 20.  We can point out that {\wm} attack has more chance of surviving under the compression process. 

\para{Watermark removing techniques.} Inpainting is a commonly used method to remove (real) watermarks. We use the inpainting method \cite{telea2004image} implemented in OpenCV \cite{OpenCVinpaint}, which required the mask of watermarks as a priori and tried to recover the watermark region according to surrounding pixels. While inpainting eliminated the watermark, because the text region overlaped with the watermark region, the inpainting method removed too many useful pixels, that is text pixels, causing OCR to fail completely.  We observe that the text even lost readability to human eyes.

\section{Conclusion and Future Work}

Generating adversarial examples for OCR systems is different from normal CV tasks.  We propose a method that successfully hides perturbations from human eyes while making them effective in the modern sequence-based OCR, by pretending perturbations as a watermark, or printing defects. We show that even with preliminary implementations, our perturbations can be still effective, transferable, and deceiving to human eyes.  

There are many future directions.  For example, allowing different watermark shapes and locations, as well as on longer sequences.  Also, it is interesting to add semantic-based (language model) attacks even further to improve the attack effectiveness. Also, the adversarial attack calls for better defense methods other than traditional image transformations.

\para{Acknowledgements.}
This work is supported in part by the National Natural Science Foundation of China (NSFC) Grant 61532001 and the Zhongguancun Haihua Institute for Frontier Information Technology.

{\small
\bibliography{egbib.bib}

\begin{thebibliography}{}

\bibitem[\protect\citeauthoryear{Bengio \bgroup et al\mbox.\egroup
  }{1995}]{bengio1995lerec}
Bengio, Y.; LeCun, Y.; Nohl, C.; and Burges, C.
\newblock 1995.
\newblock Lerec: A nn/hmm hybrid for on-line handwriting recognition.
\newblock {\em Neural Computation} 7(6):1289--1303.

\bibitem[\protect\citeauthoryear{Breuel \bgroup et al\mbox.\egroup
  }{2013}]{breuel2013high}
Breuel, T.~M.; Ul-Hasan, A.; Al-Azawi, M.~A.; and Shafait, F.
\newblock 2013.
\newblock High-performance ocr for printed english and fraktur using lstm
  networks.
\newblock In {\em 2013 12th International Conference on Document Analysis and
  Recognition},  683--687.
\newblock IEEE.

\bibitem[\protect\citeauthoryear{Brown \bgroup et al\mbox.\egroup
  }{2017}]{brown2017adversarial}
Brown, T.~B.; Man{\'e}, D.; Roy, A.; Abadi, M.; and Gilmer, J.
\newblock 2017.
\newblock Adversarial patch.
\newblock {\em arXiv preprint arXiv:1712.09665}.

\bibitem[\protect\citeauthoryear{Carlini and Wagner}{2017}]{carlini2017towards}
Carlini, N., and Wagner, D.
\newblock 2017.
\newblock Towards evaluating the robustness of neural networks.
\newblock In {\em 2017 IEEE Symposium on Security and Privacy (SP)},  39--57.
\newblock IEEE.

\bibitem[\protect\citeauthoryear{Chen \bgroup et al\mbox.\egroup
  }{2018}]{chen2018ead}
Chen, P.-Y.; Sharma, Y.; Zhang, H.; Yi, J.; and Hsieh, C.-J.
\newblock 2018.
\newblock Ead: elastic-net attacks to deep neural networks via adversarial
  examples.
\newblock In {\em Thirty-Second AAAI Conference on Artificial Intelligence}.

\bibitem[\protect\citeauthoryear{Dong \bgroup et al\mbox.\egroup
  }{2018}]{dong2018boosting}
Dong, Y.; Liao, F.; Pang, T.; Su, H.; Zhu, J.; Hu, X.; and Li, J.
\newblock 2018.
\newblock Boosting adversarial attacks with momentum.
\newblock In {\em Proceedings of the IEEE Conference on Computer Vision and
  Pattern Recognition},  9185--9193.

\bibitem[\protect\citeauthoryear{Espana-Boquera \bgroup et al\mbox.\egroup
  }{2011}]{espana2011improving}
Espana-Boquera, S.; Castro-Bleda, M.~J.; Gorbe-Moya, J.; and Zamora-Martinez,
  F.
\newblock 2011.
\newblock Improving offline handwritten text recognition with hybrid hmm/ann
  models.
\newblock {\em IEEE transactions on pattern analysis and machine intelligence}
  33(4):767--779.

\bibitem[\protect\citeauthoryear{Feinman \bgroup et al\mbox.\egroup
  }{2017}]{feinman2017detecting}
Feinman, R.; Curtin, R.~R.; Shintre, S.; and Gardner, A.~B.
\newblock 2017.
\newblock Detecting adversarial samples from artifacts.
\newblock {\em arXiv preprint arXiv:1703.00410}.

\bibitem[\protect\citeauthoryear{Goodfellow, Shlens, and
  Szegedy}{2014}]{goodfellow2014explaining}
Goodfellow, I.~J.; Shlens, J.; and Szegedy, C.
\newblock 2014.
\newblock Explaining and harnessing adversarial examples.
\newblock {\em arXiv preprint arXiv:1412.6572}.

\bibitem[\protect\citeauthoryear{Google}{2019}]{Tesseract}
Google.
\newblock 2019.
\newblock Tesseract.

\bibitem[\protect\citeauthoryear{Graves \bgroup et al\mbox.\egroup
  }{2006}]{graves2006connectionist}
Graves, A.; Fern{\'a}ndez, S.; Gomez, F.; and Schmidhuber, J.
\newblock 2006.
\newblock Connectionist temporal classification: labelling unsegmented sequence
  data with recurrent neural networks.
\newblock In {\em Proceedings of the 23rd international conference on Machine
  learning},  369--376.
\newblock ACM.

\bibitem[\protect\citeauthoryear{Grosse \bgroup et al\mbox.\egroup
  }{2017}]{grosse2017statistical}
Grosse, K.; Manoharan, P.; Papernot, N.; Backes, M.; and McDaniel, P.
\newblock 2017.
\newblock On the (statistical) detection of adversarial examples.
\newblock {\em arXiv preprint arXiv:1702.06280}.

\bibitem[\protect\citeauthoryear{Hanwei~Zhang}{2019}]{hanwei2019smooth}
Hanwei~Zhang, Yannis~Avrithis, T. F. L.~A.
\newblock 2019.
\newblock Smooth adversarial examples.
\newblock {\em arXiv preprint arXiv:1903.11862}.

\bibitem[\protect\citeauthoryear{Heng, Zhou, and
  Jiang}{2018}]{heng2018harmonic}
Heng, W.; Zhou, S.; and Jiang, T.
\newblock 2018.
\newblock Harmonic adversarial attack method.
\newblock {\em arXiv preprint arXiv:1807.10590}.

\bibitem[\protect\citeauthoryear{Hinton, Vinyals, and
  Dean}{2015}]{hinton2015distilling}
Hinton, G.; Vinyals, O.; and Dean, J.
\newblock 2015.
\newblock Distilling the knowledge in a neural network.
\newblock {\em arXiv preprint arXiv:1503.02531}.

\bibitem[\protect\citeauthoryear{Huang \bgroup et al\mbox.\egroup
  }{2017}]{huang2017densely}
Huang, G.; Liu, Z.; Van Der~Maaten, L.; and Weinberger, K.~Q.
\newblock 2017.
\newblock Densely connected convolutional networks.
\newblock In {\em Proceedings of the IEEE conference on computer vision and
  pattern recognition},  4700--4708.

\bibitem[\protect\citeauthoryear{Karmon, Zoran, and
  Goldberg}{2018}]{karmon2018lavan}
Karmon, D.; Zoran, D.; and Goldberg, Y.
\newblock 2018.
\newblock Lavan: Localized and visible adversarial noise.
\newblock {\em arXiv preprint arXiv:1801.02608}.

\bibitem[\protect\citeauthoryear{Kurakin, Goodfellow, and
  Bengio}{2016}]{kurakin2016adversarial}
Kurakin, A.; Goodfellow, I.; and Bengio, S.
\newblock 2016.
\newblock Adversarial machine learning at scale.
\newblock {\em arXiv preprint arXiv:1611.01236}.

\bibitem[\protect\citeauthoryear{Liu \bgroup et al\mbox.\egroup
  }{2016}]{liu2016delving}
Liu, Y.; Chen, X.; Liu, C.; and Song, D.
\newblock 2016.
\newblock Delving into transferable adversarial examples and black-box attacks.
\newblock {\em arXiv preprint arXiv:1611.02770}.

\bibitem[\protect\citeauthoryear{Lu, Issaranon, and
  Forsyth}{2017}]{lu2017safetynet}
Lu, J.; Issaranon, T.; and Forsyth, D.
\newblock 2017.
\newblock Safetynet: Detecting and rejecting adversarial examples robustly.
\newblock In {\em Proceedings of the IEEE International Conference on Computer
  Vision},  446--454.

\bibitem[\protect\citeauthoryear{Madry \bgroup et al\mbox.\egroup
  }{2017}]{madry2017towards}
Madry, A.; Makelov, A.; Schmidt, L.; Tsipras, D.; and Vladu, A.
\newblock 2017.
\newblock Towards deep learning models resistant to adversarial attacks.
\newblock {\em arXiv preprint arXiv:1706.06083}.

\bibitem[\protect\citeauthoryear{Moosavi-Dezfooli, Fawzi, and
  Frossard}{2016}]{moosavi2016deepfool}
Moosavi-Dezfooli, S.-M.; Fawzi, A.; and Frossard, P.
\newblock 2016.
\newblock Deepfool: a simple and accurate method to fool deep neural networks.
\newblock In {\em Proceedings of the IEEE Conference on Computer Vision and
  Pattern Recognition},  2574--2582.

\bibitem[\protect\citeauthoryear{Nguyen, Yosinski, and
  Clune}{2015}]{nguyen2015deep}
Nguyen, A.; Yosinski, J.; and Clune, J.
\newblock 2015.
\newblock Deep neural networks are easily fooled: High confidence predictions
  for unrecognizable images.
\newblock In {\em Proceedings of the IEEE conference on computer vision and
  pattern recognition},  427--436.

\bibitem[\protect\citeauthoryear{OpenCV}{a}]{OpenCVinpaint}
OpenCV.
\newblock inpainting(cv.inpaint).
\newblock
  \url{https://docs.opencv.org/3.0-beta/modules/photo/doc/inpainting.html}.

\bibitem[\protect\citeauthoryear{OpenCV}{b}]{OpenCVweb}
OpenCV.
\newblock Median filter(cv.medianblur).
\newblock
  \url{https://docs.opencv.org/master/d4/d13/tutorial_py_filtering.html}.

\bibitem[\protect\citeauthoryear{Papernot \bgroup et al\mbox.\egroup
  }{2016a}]{papernot2016limitations}
Papernot, N.; McDaniel, P.; Jha, S.; Fredrikson, M.; Celik, Z.~B.; and Swami,
  A.
\newblock 2016a.
\newblock The limitations of deep learning in adversarial settings.
\newblock In {\em 2016 IEEE European Symposium on Security and Privacy
  (EuroS\&P)},  372--387.
\newblock IEEE.

\bibitem[\protect\citeauthoryear{Papernot \bgroup et al\mbox.\egroup
  }{2016b}]{papernot2016distillation}
Papernot, N.; McDaniel, P.; Wu, X.; Jha, S.; and Swami, A.
\newblock 2016b.
\newblock Distillation as a defense to adversarial perturbations against deep
  neural networks.
\newblock In {\em 2016 IEEE Symposium on Security and Privacy (SP)},  582--597.
\newblock IEEE.

\bibitem[\protect\citeauthoryear{Papernot \bgroup et al\mbox.\egroup
  }{2017}]{papernot2017practical}
Papernot, N.; McDaniel, P.; Goodfellow, I.; Jha, S.; Celik, Z.~B.; and Swami,
  A.
\newblock 2017.
\newblock Practical black-box attacks against machine learning.
\newblock In {\em Proceedings of the 2017 ACM on Asia Conference on Computer
  and Communications Security},  506--519.
\newblock ACM.

\bibitem[\protect\citeauthoryear{Papernot, McDaniel, and
  Goodfellow}{2016}]{papernot2016transferability}
Papernot, N.; McDaniel, P.; and Goodfellow, I.
\newblock 2016.
\newblock Transferability in machine learning: from phenomena to black-box
  attacks using adversarial samples.
\newblock {\em arXiv preprint arXiv:1605.07277}.

\bibitem[\protect\citeauthoryear{Sharma and Chen}{2017}]{sharma2017attacking}
Sharma, Y., and Chen, P.-Y.
\newblock 2017.
\newblock Attacking the madry defense model with $ l\_1 $-based adversarial
  examples.
\newblock {\em arXiv preprint arXiv:1710.10733}.

\bibitem[\protect\citeauthoryear{Smith}{2007}]{smith2007overview}
Smith, R.
\newblock 2007.
\newblock An overview of the tesseract ocr engine.
\newblock In {\em Ninth International Conference on Document Analysis and
  Recognition (ICDAR 2007)}, volume~2,  629--633.
\newblock IEEE.

\bibitem[\protect\citeauthoryear{Szegedy \bgroup et al\mbox.\egroup
  }{2013}]{szegedy2013intriguing}
Szegedy, C.; Zaremba, W.; Sutskever, I.; Bruna, J.; Erhan, D.; Goodfellow, I.;
  and Fergus, R.
\newblock 2013.
\newblock Intriguing properties of neural networks.
\newblock {\em arXiv preprint arXiv:1312.6199}.

\bibitem[\protect\citeauthoryear{Telea}{2004}]{telea2004image}
Telea, A.
\newblock 2004.
\newblock An image inpainting technique based on the fast marching method.
\newblock {\em Journal of graphics tools} 9(1):23--34.

\bibitem[\protect\citeauthoryear{Tram{\`e}r \bgroup et al\mbox.\egroup
  }{2017}]{tramer2017ensemble}
Tram{\`e}r, F.; Kurakin, A.; Papernot, N.; Goodfellow, I.; Boneh, D.; and
  McDaniel, P.
\newblock 2017.
\newblock Ensemble adversarial training: Attacks and defenses.
\newblock {\em arXiv preprint arXiv:1705.07204}.

\bibitem[\protect\citeauthoryear{Wang \bgroup et al\mbox.\egroup
  }{2012}]{wang2012end}
Wang, T.; Wu, D.~J.; Coates, A.; and Ng, A.~Y.
\newblock 2012.
\newblock End-to-end text recognition with convolutional neural networks.
\newblock In {\em Proceedings of the 21st International Conference on Pattern
  Recognition (ICPR2012)},  3304--3308.
\newblock IEEE.

\bibitem[\protect\citeauthoryear{Xu, Evans, and Qi}{2017}]{Xu2017Feature}
Xu, W.; Evans, D.; and Qi, Y.
\newblock 2017.
\newblock Feature squeezing: Detecting adversarial examples in deep neural
  networks.

\end{thebibliography}
\bibliographystyle{aaai}
}

\end{document}